\documentclass{article}
\usepackage{amsmath,graphicx,mlspconf,color}
\usepackage[section]{placeins}
\usepackage{booktabs}
\usepackage{multirow}
\usepackage{bbding}
\usepackage[marginal]{footmisc}
\usepackage{amsmath,amssymb,amsfonts}
\usepackage{url}

\toappear{2024 IEEE International Workshop on Machine Learning for Signal Processing, Sept.\ 22--25, 2024, London, UK}

\title{MULTIMODAL GENERATIVE SEMANTIC COMMUNICATION BASED ON LATENT DIFFUSION MODEL}
\name{%
    Weiqi Fu$^{\star}$%
    \qquad Lianming Xu$^{\dagger}$%
    \qquad Xin Wu$^{\star}$%
    \qquad Haoyang Wei$^{\star}$%
    \qquad Li Wang$^{\star}$\thanks{This work was supported in part by the National Natural Science Foundation of China under grants U2066201, 62171054, 62101045, and 62201071, in part by the Natural Science Foundation of Beijing Municipality under Grant L222041, in part by the Fundamental Research Funds for the Central Universities under Grant No. 24820232023YQTD01, No. 2023RC96, and No. 2024RC06, in part by the Double First-Class Interdisciplinary Team Project Funds 2023SYLTD06. (Corresponding author: Li Wang)}%
}
\address{%
    $^{\star}$ School of Computer Science (National Pilot Software Engineering School), \\Beijing University of Posts and Telecommunications, Beijing, China \\%
    $^{\dagger}$ School of Electronic Engineering, Beijing University of Posts and Telecommunications, Beijing, China \\%
    Email:{\{fuweiqi, xulianming, xin.wu, liwang\}@bupt.edu.cn, why22461@gmail.com}
}

\begin{document}

\maketitle

\begin{abstract}
In emergencies, the ability to quickly and accurately gather environmental data and command information, and to make timely decisions, is particularly critical. Traditional semantic communication frameworks, primarily based on a single modality, are susceptible to complex environments and lighting conditions, thereby limiting decision accuracy. To this end, this paper introduces a multimodal generative semantic communication framework named mm-GESCO. The framework ingests streams of visible and infrared modal image data, generates fused semantic segmentation maps, and transmits them using a combination of one-hot encoding and zlib compression techniques to enhance data transmission efficiency. At the receiving end, the framework can reconstruct the original multimodal images based on the semantic maps. Additionally, a latent diffusion model based on contrastive learning is designed to align different modal data within the latent space, allowing mm-GESCO to reconstruct latent features of any modality presented at the input. Experimental results demonstrate that mm-GESCO achieves a compression ratio of up to 200 times, surpassing the performance of existing semantic communication frameworks and exhibiting excellent performance in downstream tasks such as object classification and detection.
\end{abstract}

\section{Introduction}
\label{sec:intro}
The timely acquisition of on-site disaster situation awareness is crucial for the success of emergency rescue operations. Deploying drones to disaster sites to collect and transmit data to the command center has become standard practice. However, complex environments, such as densely forested areas with tree obstructions, can hinder the rapid location of individuals needing rescue when information is captured using only visible light devices. Integrating additional sensing devices, such as infrared sensors, is essential to enhance the sensing capabilities of drones in such demanding environments.

\begin{figure*}[!t]
    \centering
    \includegraphics[width=1.0\textwidth]{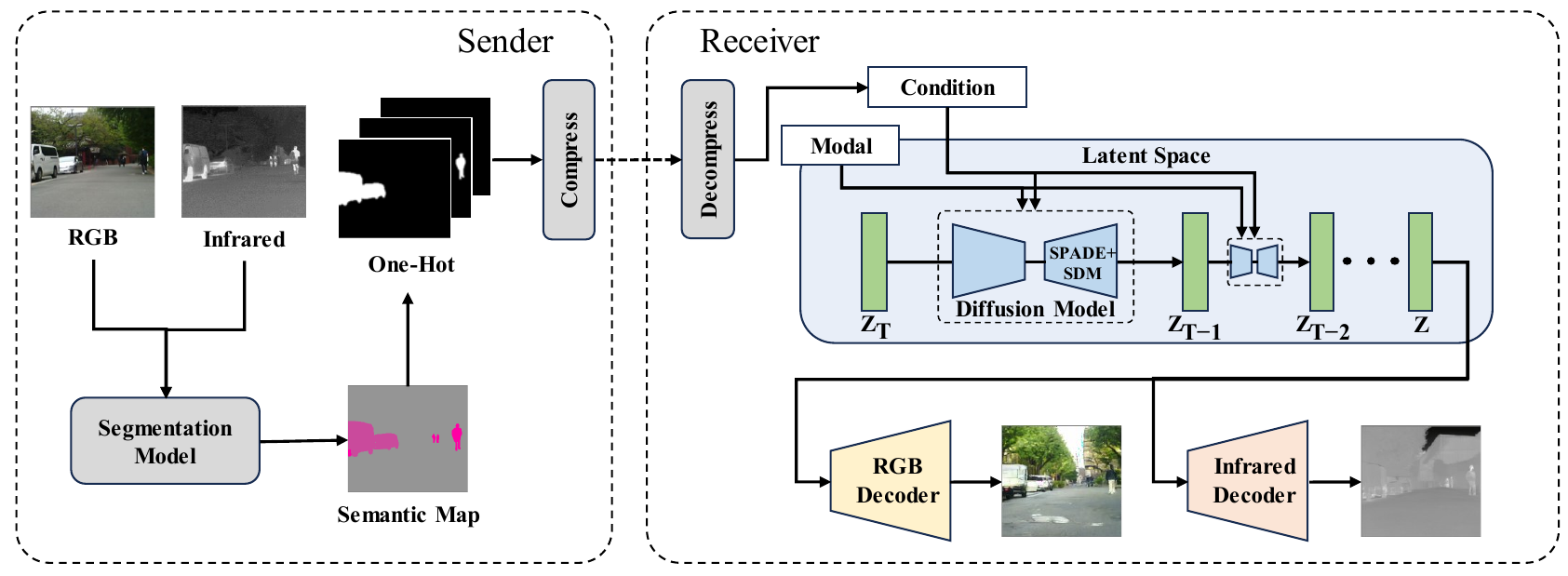}
    \caption{The multi-modal generative semantic communication framework}
    \label{fig01}
\end{figure*}

Recent advances in drone technology, including increased payload capacities and extended flight durations, support the incorporation of multiple sensor modalities. However, the use of multimodal sensing devices entails transmitting a larger volume of sensory data back to the command center. When disasters occur, local infrastructure is typically destroyed, resulting in a lack of public network support. Consequently, drones must rely on temporarily deployed dedicated networks for data transmission. The complexity of the disaster environment and the extended distance between drones and the command center create a long transmission link. This situation results in limited communication bandwidth and unstable transmission links, making it extremely difficult to transmit complete raw sensory data, especially when collected through multi-modal devices.

Inspired by deep learning (DL) technologies, DL-based frameworks prioritize the extraction and transmission of key semantic information, which include texts \cite{c1}, images \cite{c2}, and speech \cite{c3}. This approach significantly reduces the bandwidth requirements for data transmission. Furthermore, the development of generative AI models, such as Variational Autoencoders (VAE), Generative Adversarial Networks (GAN), and denoising diffusion models \cite{c6}, has enabled the reconstruction of original data at the receiver based on the transmitted semantic data \cite{c13,c14,c7}. This reconstructed data can be applied to downstream tasks such as image classification, depth estimation, and other related applications. By integrating semantic communication with downstream tasks \cite{c8}, it is possible to enhance data processing performance. 

Existing studies primarily focused on the reconstruction of data within single-modal, which limits the applicability in more complex scenarios with multiple tasks. Addressing these limitations, recent research has made significant advancements. In \cite{c9}, a deep neural network-enabled semantic communication framework was proposed to execute the visual question-answering task. It supports both image and text modalities. In \cite{c10}, a unified semantic communication framework was proposed, capable of handling multiple tasks with multiple modalities of data using a single fixed model. Further research is still needed on multi-modal generative semantic communication.

The main contributions of this paper can be summarized as follows.
\begin{itemize}
\item We proposed a multi-modal generative semantic communication framework, named mm-GESCO, which extracts and transmits semantic segmentation maps fused from visible light and infrared images. At the receiver, it reconstructs the data into visible light and infrared images based on the semantic segmentation maps. By employing one-hot encoding and zlib compression, we achieved nearly 200x compression of a single semantic segmentation map, significantly enhancing the success rate of sensory data back-haul in emergency environments.
\item We introduced the latent diffusion model, which utilizes contrastive learning methods alongside a pair of autoencoders to align data from visible light and infrared modalities within the latent space. By using modal information as the condition for the diffusion model, we enabled a single model to reconstruct data across various modalities, effectively reducing the deployment overhead in emergencies.

\item Experiments demonstrate that our proposed mm-GESCO outperforms existing semantic communication frameworks, whether single or multiple modalities, achieving superior performance in downstream tasks such as object classification and detection.
\end{itemize}

\section{problem Description}

In this paper, we explore how to efficiently transmit multi-modal images, specifically visible light and infrared modalities, through a semantic communication framework and subsequently reconstruct these modalities. Given the limited transmission resources in emergency scenarios, the data transmitted by the sender should be minimized in size. Furthermore, in back-haul scenarios, the sender is generally constrained by limited computational resources, which is in contrast to the receiver who usually has substantial computational capabilities. A lightweight, well-trained model should be deployed on the sender to alleviate the computational overhead during the inference process.

For the sender, given that the UAV concurrently captures the disparate modal images at identical temporal and spatial coordinates, it is feasible to employ a multi-modal semantic segmentation model to fuse the semantic information from various modal images. Only one semantic segmentation map needs to be transmitted, obviating the necessity to extract individual semantic segmentation maps for each modality. Additionally, the integration of one-hot encoding and compression techniques can be utilized to reduce the size of the semantic segmentation map. As the volume of transmission data is minimized, the strategic application of channel coding techniques will significantly enhance the success rate of data back-haul in emergency scenarios. In this paper, we exclude the impact of transmission noise, which is commonly addressed by a separate channel encoder. We concentrate on semantic coding to achieve efficient data compression, prioritizing the reduction of transmitted data size and its high-fidelity reconstruction at the receiver.

For the receiver, to reconstruct the original modalities from the fused semantic data, we introduce a multi-modal diffusion model to reconstruct data across various modalities. Considering that different modalities share one identical semantic segmentation map, the mere use of the one-hot encoded semantic segmentation map as a conditional input for the diffusion model is inadequate. Consequently, we incorporate modality categories as conditional inputs into the diffusion model. Furthermore, we implement a latent diffusion model (LDM) \cite{c11} and integrate contrastive learning to train autoencoders for both visible light and infrared modalities concurrently. This approach facilitates the alignment of features across modalities within the latent space, effectively minimizing the disparities between the features generated by the diffusion model for different modalities, thereby enhancing the multi-modal generation capabilities of the diffusion model.

To evaluate the performance of the proposed framework, mm-GESCO, we employ not only traditional metrics such as Learned Perceptual Image Patch Similarity (LPIPS) \cite{LPIPS} and Fréchet Inception Distance (FID) \cite{FID}, but also consider object classification and detection as downstream tasks, given that the most critical task in emergency scenarios is search and rescue. Therefore, in addition to LPIPS and FID, the performance evaluation of the generative semantic communication framework is conducted by comparing the object classification and detection outcomes on images before and after reconstruction.

\begin{figure}[!t]
    \centering
    \includegraphics[width = 0.49\textwidth]{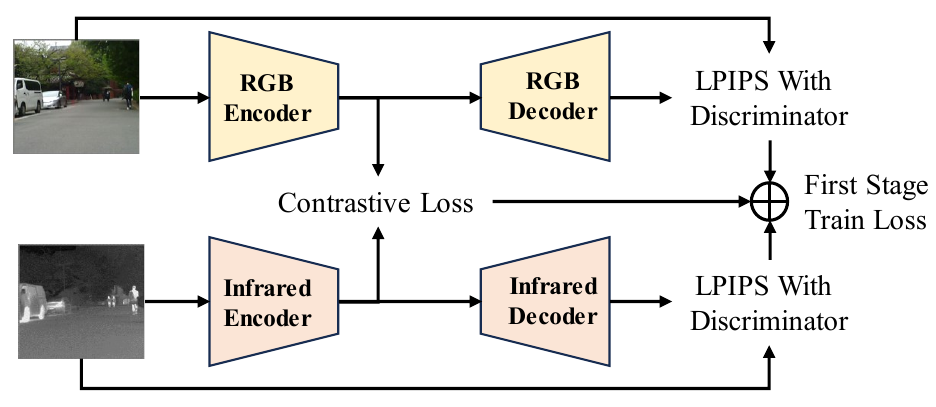}
    \caption{Training the autoencoders}
    \label{fig02}
\vspace{-0cm}
\end{figure}

\section{Proposed framework}

Building upon the framework established in previous studies \cite{c7}, this paper introduces a novel multimodal generative semantic communication framework, named mm-GESCO, specifically designed for deployment in emergency situations. Given the critical importance of processing visible light and infrared multimodal data, the proposed framework addresses the urgent need that not only withstands severe communication constraints but also operates effectively under the limited computational resources typical of emergency environments.

\subsection{Multimodal Generative Semantic Communication}

As shown in Fig.~\ref{fig01}, the overall framework consists of two parts: the sender and the receiver. At the sender, we first employ a multimodal semantic segmentation model that effectively extracts fused semantic information from various modalities. Then, to adapt to the extreme communication conditions prevalent in emergency scenarios, we integrate compression algorithms to achieve high compression rates of semantic information without loss. At the receiver, we utilize a multimodal diffusion model to generate semantically consistent data across various modalities. Through this framework, we provide semantically consistent visual results for rescue personnel and high-quality data for downstream tasks such as object classification and detection.

\textbf{Sender.} Our framework begins with the application of a semantic segmentation model to extract essential information from the multi-modal data. We apply MFNet \cite{c12}, which is designed for visible light and infrared data and is also lightweight, making it suitable for deployment on emergency devices. After processing through MFNet, the fused semantic segmentation map is first encoded using one-hot coding. Conditioning the receiver's diffusion model with this one-hot encoded data has demonstrated improved performance. This representation is then transformed into binary format and compressed using zlib, which can achieve significant compression, potentially up to 200x based on our actual tests. Due to the effective compression, it will be easier to enhance the reliability of data transmission under extreme communication conditions by combining channel coding and other technologies. 

\textbf{Receiver.} The core of our framework is a latent diffusion model. This model begins with a noise sample ($x_0 \sim N(0, I)$) and progressively removes noise to generate features in the latent space. As the reconstruction is carried out within this latent space, both the training and sampling efficiency can be significantly improved without degrading the quality of the outputs. Additionally, the two-stage training mechanism of the latent diffusion model provides a practical way to extend to multi-modal applications. The process starts with training an autoencoder, where the encoder maps the original image into the latent space, and the decoder reconstructs the image from the latent space features. Following this, a denoising diffusion model is trained to reconstruct these latent space features, using the encoded features as ground truth. In this paper, to achieve multi-modal reconstruction, each modality is associated with a dedicated autoencoder. Furthermore, this latent diffusion model is conditioned on both semantic segmentation maps and modality categories.

\subsection{First Stage: Training the Autoencoders}

Before training our diffusion model, we initially focuse on training autoencoders for each specific modality. The encoders within these autoencoders are designed to project the data from each modality onto a unified latent space. As shown in Fig.~\ref{fig02}, we incorporate contrastive loss into our training procedure. Therefore. the loss function of the first stage of training is defined as follows: 
\begin{equation}
\mathcal{L}_{AE} = \mathcal{L}_{C} + \mathcal{L}_{RGB} + \mathcal{L}_{Infrared},
\end{equation}
where $\mathcal{L}_{C}$ is the contrastive loss. And both $\mathcal{L}_{RGB}$ and $\mathcal{L}_{Infrared}$ are a combination of a perceptual loss and a patch-based adversarial objective, which are utilized in the training process of LDM \cite{c11}. By encoding the data from visible light and infrared into a shared latent space through their respective encoders, we optimize the diffusion model for handling diverse modalities while maintaining coherence in the generated outputs.

\subsection{Second Stage: Training the Diffusion Model in Latent Space}
\begin{figure}[!t]
    \centering
    \includegraphics[width = 0.49\textwidth]{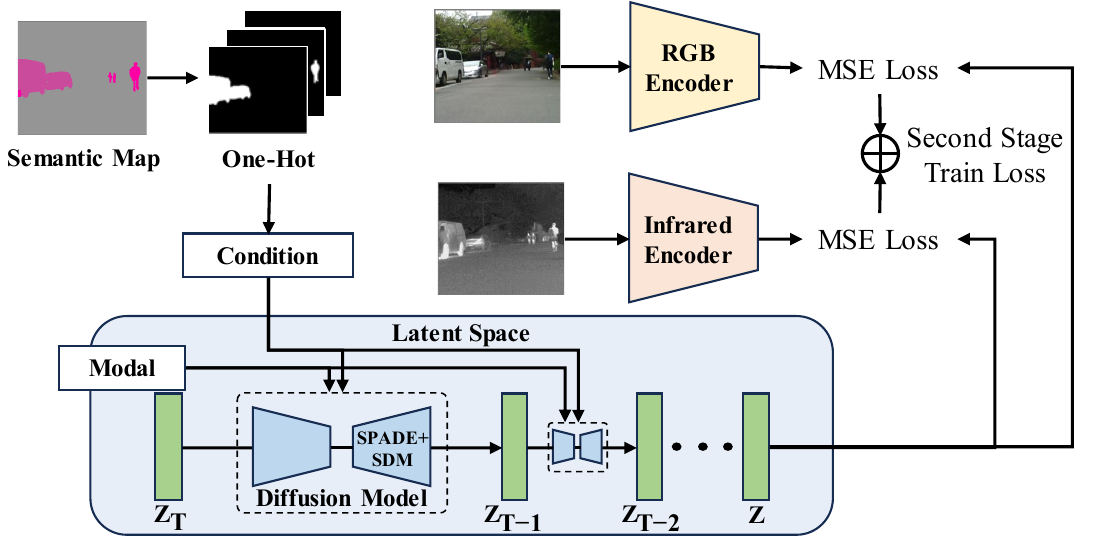}
    \caption{Training the diffusion model in latent space}
    \label{fig03}
\vspace{-0cm}
\end{figure}

Although we introduce contrastive loss in the autoencoders to align features across different modalities as much as possible, generating multimodal data with a single diffusion model still poses a challenge and requires further improvement. As shown in Fig.~\ref{fig03}, we incorporate both one-hot encoded semantic segmentation map and modality categories as the condition of the diffusion model, so that our model can generate latent space features for each modality separately. This is essential because both visible light and infrared modalities use the same semantic segmentation map due to the multi-modal fused segmentation model. To ensure the reconstruction accuracy and consistency across modalities, we compute the mean squared error (MSE) loss for the generated latent space features of each modality against their corresponding ground truth, which is derived from encoding the original image using the encoder parts of autoencoders. These MSE losses are then summed to optimize the model's performance, ensuring that it accurately minimizes discrepancies across both visible light and infrared images. which can be expressed as follows:

\begin{equation}
\begin{split}\mathcal{L}_{LDM} = MSE(\epsilon_0(x_0),\theta(x,t,y,0)) \\
+ MSE(\epsilon_1(x_1),\theta(x,t,y,1)) \end{split},
\end{equation}
where $\epsilon$ outputs the feature encoded by the autoencoder specific to each modality. The reconstruction result of the diffusion model, denoted by $\theta$, is conditioned on both the one-hot encoded data $y$ and the modality categories. Specifically, 0 represents visible light data and 1 represents infrared data. This approach enhances the model's ability to handle multi-modal data effectively.

\section{Experimental Evaluation}
In this section, we present the results of comparative experiments and ablation studies that assess the performance of our proposed mm-GESCO framework. We begin by outlining the experimental setup and the criteria used for evaluation. Then we detail the outcomes of the experiments.

\begin{figure*}[t]
    \centering
    \includegraphics[width=1\textwidth]{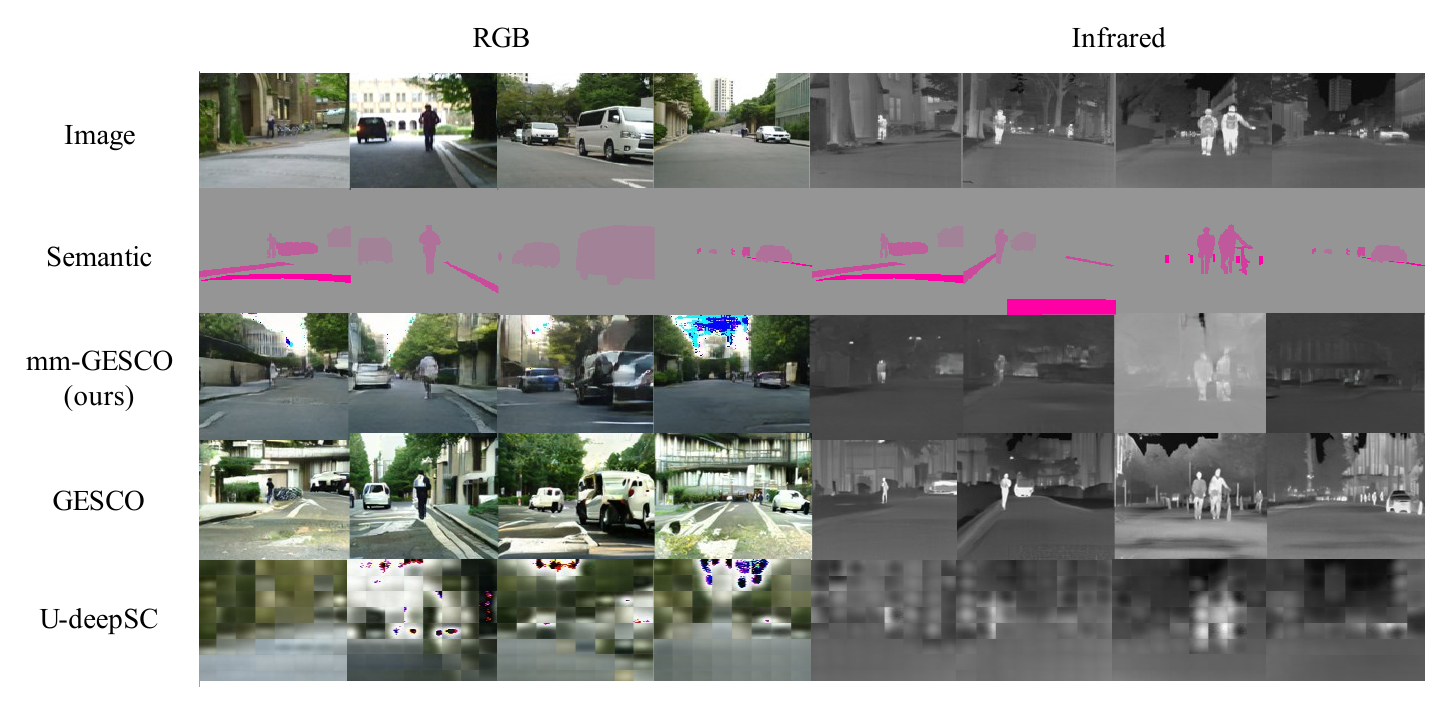}
    \caption{Image reconstruction examples of mm-GESCO(ours), GESCO, and U-deepSC}
    \label{fig04}
    \vspace{-0.3cm}
\end{figure*}

\subsection{Setup}
In this paper, we use \textit{Multi-spectral Semantic Segmentation Dataset}\footnote{\url{https://www.mi.t.u-tokyo.ac.jp/static/projects/mil_multispectral}} published by MFNet, which comprises 1,569 urban street images, including visible light and thermal infrared images. Originally designed for semantic segmentation tasks, this dataset provides semantic segmentation maps as ground truth, annotated with eight classes of common urban street obstacles (car, person, bike, curve, car stop, guardrail, color cone, and bump). 

\begin{table}[h!t]
\centering
\caption{\centering Performance comparison between mm-GESCO, U-deepSC, and GESCO}
\resizebox{1\columnwidth}{!}
{
\tiny
\renewcommand{\arraystretch}{1.7}
\setlength{\tabcolsep}{3pt}
\begin{tabular}{c|c|cccc|c}
\toprule 
\multicolumn{2}{c|}{\multirow{2}{*}{\textbf{Framework}}}&\multirow{2}{*}{\textbf{FID}}&\multirow{2}{*}{\textbf{LPIPS}}&{\textbf{Classification}}&{\textbf{Detection}}&\textbf{Transmit} \\
\multicolumn{2}{c|}{}&&&\textbf{Accuracy}&\textbf{mIoU}&\textbf{data size} \\
\hline
\multirow{3}{*}{\textbf{RGB}}&{\textbf{mm-GESCO}}&\textbf{85}&0.62&\textbf{68\%}&\textbf{0.060}&\multirow{6}{*}{\parbox{15pt}{ \centering About \\ 700 \\ bytes}} \\
\cline{2-6}
&{\textbf{U-deepSC}}&301&\textbf{0.48}&63\%&0.000 \\
\cline{2-6}
&{\textbf{GESCO}}&129&0.68&63\%&0.035 \\
\cline{0-5}
\multirow{3}{*}{\textbf{Infrared}}&{\textbf{mm-GESCO}}&\textbf{119}&0.56&60\%&\textbf{0.089} \\
\cline{2-6}
&{\textbf{U-deepSC}}&295&\textbf{0.35}&52\%&0.000 \\
\cline{2-6}
&{\textbf{GESCO}}&128&0.50&\textbf{77\%}&0.058 \\
\bottomrule 
\end{tabular}
}
\label{tab01}
\vspace{-0cm}
\end{table}

To evaluate the performance of our proposed framework, we employed the FID and LPIPS metrics in our image reconstruction task. The FID score correlates with human judgment, while the LPIPS score measures perceptual similarity. Lower values in both metrics signify a greater similarity between the generated and original images, indicating more effective image reconstruction. After assessing image reconstruction, we evaluated the framework's performance on the tasks of object classification and detection. We employed a pretrained Yolov8 model to detect pedestrians in the images both before and after reconstruction. The effectiveness of our method was quantified by calculating the accuracy of the classification labels and the mean Intersection over Union (mIoU) of the bounding boxes around detected pedestrians. 

To compare the effectiveness of our method with the single-modal generative semantic communication method (GESCO) and the multimodal semantic communication method (U-deepSC), we conducted experiments on the image reconstruction, classification and detection tasks, using images resized to $128\times 128$ pixels. The analysis was performed on a single RTX 4090 graphics card.

\subsection{Comparisons and Results Analysis}
Table \ref{tab01} compares the performance of three methods in visible light image reconstruction and infrared image reconstruction, respectively. As our primary goal is the multimodality of semantic communication, the experiment did not account for transmission noise. Empirical data from compressing using the zlib technique shows that the mean file size of the $128\times 128$ semantic segmentation maps required for transmission in both our method and GESCO approximates 700 bytes per image. To maintain this transmission size, we modified the configuration of the channel encoding module within the U-deepSC method, ensuring that the size of the output features to be transmitted closely matches that of a single compressed semantic segmentation map.

The experimental results indicate that our method achieved better performance in terms of FID score and the downstream tasks of classification and detection. Despite being compared with single-modal methods, our multi-modal method benefits from its innovative design in the latent space. In contrast, the U-deepSC method demonstrated a comparative advantage in terms of LPIPS score, likely due to its semantic communication based on encoders and decoders. This setup enables the transmission of relatively complete feature information, thus leading to a better LPIPS score. Our method transmits only partial semantic features of objects, making these features more prominent in the generated images and resulting in better performance in downstream tasks.

\subsection{Ablation Study}
\vspace{-0.3cm}
\begin{table}[h!t]
\centering
\caption{Ablation experiment of mm-GESCO}
\resizebox{1\columnwidth}{!}
{
\renewcommand{\arraystretch}{1.5}
\setlength{\tabcolsep}{3pt}
\small
\begin{tabular}{c|c|c|cccc}
\toprule 
&\textbf{Contrastive}&\textbf{Modality as}&\multirow{2}{*}{\textbf{FID}}&\multirow{2}{*}{\textbf{LPIPS}}&{\textbf{Classification}}&{\textbf{Detection}} \\
&\textbf{loss}&\textbf{condition}&&&\textbf{Accuracy}&\textbf{mIoU} \\
\hline
\multirow{3}{*}{\textbf{RGB}}&&&113&0.66&62\%&0.023 \\
\cline{2-7}
&\Checkmark&&96&0.63&65\%&0.046 \\
\cline{2-7}
&\Checkmark&\Checkmark&\textbf{85}&\textbf{0.62}&\textbf{68\%}&\textbf{0.060} \\
\hline
\multirow{3}{*}{\textbf{Infrared}}&&&136&0.60&56\%&0.053 \\
\cline{2-7}
&\Checkmark&&119&0.57&59\%&0.061 \\
\cline{2-7}
&\Checkmark&\Checkmark&\textbf{119}&\textbf{0.56}&\textbf{60\%}&\textbf{0.089} \\

\bottomrule 
\end{tabular}
}
\label{tab02}
\vspace{-0cm}
\end{table}

As shown in Table \ref{tab02}, this paper conducted ablation experiments on our method. We investigated two scenarios: whether training the autoencoder with an additional contrastive loss and whether including modality categories as another condition input to the diffusion model. Comparative experimental results were provided. By comparing the FID score, LPIPS score, and the results of downstream tasks, it is evident that adding the contrastive loss and including modality categories as another condition both improved performance.

\section{Conclusions}
In this paper, we propose the mm-GESCO framework. At the sender, we fuse the multi-modal images, such as visible light and infrared. We extract the semantic segmentation maps and combine them with compression coding and channel coding techniques, effectively reducing bandwidth demands and improving the success rate of data back-haul. At the receiver, the framework utilizes a latent diffusion model to reconstruct images from various modalities, ensuring the accurate transmission of command information and facilitating better integration with downstream tasks. 
Experimental results indicate that our framework achieves superior reconstruction performance compared to existing single-modal and multi-modal semantic communication frameworks. In future work, we will refine the mm-GESCO framework to expand its applicability to additional modalities.

\bibliographystyle{IEEEbib}
\bibliography{strings,refs}

\end{document}